\documentclass{article}

\usepackage[final]{nips_2018}

\usepackage[utf8]{inputenc}
\usepackage[english]{babel}
\usepackage[T1]{fontenc}

\usepackage{amsfonts}
\usepackage{amsmath}
\usepackage{amssymb}
\usepackage{amsthm}
\usepackage{enumitem}
\usepackage{bigints}
\usepackage{booktabs}
\usepackage{hyperref}
\usepackage{algorithm}
\usepackage{algorithmic}
\usepackage[disable]{todonotes}

\setcitestyle{authoryear, round}

\usepackage{subcaption}

\usepackage{mycommands}
\theoremstyle{plain}

\newtheorem*{bound*}{Code length}

\newcommand{\deq}{\mathrel{\mathop:}=}

\newcommand{\optional}[1]{#1}   

\title{The Description Length of Deep Learning Models}

\author{
  Léonard Blier\\
  École Normale Supérieure\\
  Paris, France\\
  \texttt{leonard.blier@normalesup.org}\\
  \And
  Yann Ollivier\\
  Facebook Artificial Intelligence Research\\
  Paris, France\\
  \texttt{yol@fb.com}
}

\begin{document}
\maketitle

\begin{abstract}
Solomonoff's general theory of inference \citep{Solomonoff1964} and the
Minimum Description Length principle \citep{Grunwald, rissanen2007information} formalize
Occam's razor, and hold that a good model of data
is a model that is good at losslessly compressing the data, including the cost of
describing the model itself. 
Deep neural networks might seem to go against this principle given the large number of parameters to be encoded.

We demonstrate experimentally the ability of deep neural
networks to compress the training data even when accounting
for 
parameter encoding. The compression viewpoint
originally motivated the use of
\emph{variational methods} in neural networks
\citep{Hinton,Schmidhuber1997}. Unexpectedly, we
found that these variational methods provide surprisingly poor compression
bounds, despite being explicitly built to minimize such bounds.
This might explain the relatively poor practical
performance of variational methods in deep learning.
On the other hand,
simple incremental
encoding methods
yield excellent compression values on deep networks, vindicating
Solomonoff's approach.
\end{abstract}

\section{Introduction}
\label{sec:introduction}

Deep learning has achieved remarkable results in many different areas
\citep{LeCun2015}. Still, the ability of deep models not to overfit
despite their large number of parameters is not well understood. To
quantify
the complexity of these models in light of their
generalization ability, several metrics beyond parameter-counting have
been measured, such as the number of degrees of freedom of models
\citep{Gao2016}, or their intrinsic dimension \citep{Li2018}. These
works concluded that deep learning models are significantly simpler than
their numbers of parameters might suggest.



In information theory and Minimum Description Length (MDL), learning a
good model of the data is recast as using the model to losslessly
transmit the data in as few bits as possible.  More complex models
will compress the data more, but the model must be transmitted as well.
The overall codelength can be understood as a
combination of quality-of-fit of the model (compressed data length), together with the cost of
encoding (transmitting) the model itself.
For neural networks, the MDL viewpoint goes back as far as
\citep{Hinton}, which used a variational technique to estimate the joint
compressed length of data and parameters in a neural network model.

Compression is strongly related to generalization and practical
performance. Standard sample complexity bounds (VC-dimension,
PAC-Bayes...) are related to the compressed length of the data in a
model, and any compression scheme leads to generalization bounds
\citep{Blum2003}. Specifically for deep learning, \citep{Arora} showed that
compression leads to generalization bounds (see also \citep{Dziugaite2017}). Several other deep learning
methods have been inspired by information theory and the compression
viewpoint. In unsupervised learning, autoencoders and especially
variational autoencoders \citep{Kingma2013} are compression methods of the
data \citep{Ollivier2014a}. In supervised learning, the
information bottleneck method studies how the hidden representations in a
neural network compress the inputs while preserving the mutual information
between inputs and outputs
\citep{Tishby2015,Shwartz-Ziv,Achille2017}.

MDL is based on Occam's razor, and on Chaitin's hypothesis that
``\emph{comprehension is compression}'' \citep{Chaitin2002}: any
regularity in the data can be exploited both to compress it and to make
predictions.  This is ultimately rooted in Solomonoff's general theory of
inference \citep{Solomonoff1964} (see also, e.g.,
\citep{Hutter2007,Schmidhuber1997}), whose principle is to favor
models that correspond to the ``shortest program'' to produce the
training data, based on its Kolmogorov complexity \citep{Li2008a}. If no
structure is present in the data, no compression to a shorter program is
possible.

\begin{figure}[t]
  \centering
  \includegraphics[width=.7\linewidth]
{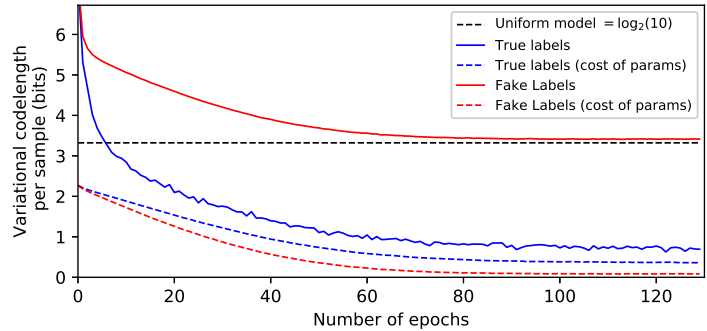}
  \caption{\textbf{Fake labels cannot be compressed} Measuring codelength
  while training a deep
  model on MNIST with true and fake labels.
  The model is an MLP with 3 hidden layers of size 200, with RELU units.
  With ordinary SGD training, the model is able to overfit random labels.
  The plot shows the effect of using variational learning instead, and
  reports the variational objective (encoding cost of the training data,
  see Section \ref{sec:varcode}),
  on true and fake labels. We also isolated the
  contribution from parameter encoding in the total loss (KL term in
  \eqref{eq:boundvar}). With true labels, the encoding cost is below the
  uniform encoding, and half of the description length is information
  contained in the weights. With fake labels, on the contrary, the
  encoding cost converges to a uniform random model, with no information contained in the weights: there is no mutual information between inputs and outputs.}
  \label{fig:fakevi}
\end{figure}

The problem of overfitting fake
labels is a nice illustration: convolutional neural networks commonly
used for image classification are able to reach $100\%$ accuracy on
random labels on the train set \citep{Zhang2016}.  However, measuring the
associated compression bound (Fig.~\ref{fig:fakevi}) immediately reveals that these models do not
\emph{compress} fake labels (and indeed, theoretically, they cannot, see
Appendix~\ref{app:fake-noncompressible}), that no information is present
in the model parameters, and that no learning has
occurred.

In this work we explicitly measure how much
current deep models actually compress data. 
\optional{(We introduce no new architectures or learning procedures.)}
As seen above,
this may clarify several issues around generalization and measures of model complexity. Our contributions are: 
\begin{itemize}
\item We show that the traditional method to estimate MDL codelengths in deep
learning, variational inference \citep{Hinton}, yields surprisingly
inefficient codelengths for deep models, despite explicitly minimizing
this criterion. This might explain why variational inference as a
regularization method often does not
reach optimal test performance.


\item We introduce new practical ways to compute tight compression bounds
in deep learning models, based on the MDL toolbox \citep{Grunwald, rissanen2007information}.
We show that \emph{prequential coding} on
top of standard learning,
yields much better codelengths than variational inference, correlating better with test set performance.
Thus, despite their many
parameters, deep learning models do compress the data well, even when
accounting for the cost of describing the model. 
\end{itemize}

\section{Probabilistic Models, Compression, and Information Theory}

Imagine that Alice
wants to efficiently transmit some information to Bob. Alice has a dataset $\mathcal{D} =
\{(x_{1}, y_{1}), ..., (x_{n}, y_{n})\}$ where $x_{1}, ..., x_{n}$ are
some inputs and $y_{1}, ..., y_{n}$ some labels. We do not assume that
these data come from a ``true'' probability distribution. Bob also has the data
$x_{1}, ..., x_{n}$, but he does not have the labels. This describes a
\emph{supervised learning} situation in which the inputs $x$ may be
publicly available, and a prediction of the labels $y$ is needed. How
can deep learning models help with data encoding? One key problem is that Bob
does not necessarily know the precise, trained model that Alice is using.
So some explicit or implicit transmission of the model itself is required.

We study, in turn, various methods to encode the labels $y$, with or
without a deep learning model. Encoding the labels knowing the inputs is
equivalent to estimating their mutual information (Section~\ref{sec:mi});
this is distinct from the problem of practical network compression
(Section~\ref{sec:netcompression}) or from using neural networks for lossy data
compression.
Our running example will be image classification on the MNIST
\citep{lecun1988} and CIFAR10 \citep{Krizhevsky2009} datasets.

\subsection{Definitions and notation}
\label{sec:notation}

Let
$\mathcal{X}$ be the input space and $\mathcal{Y}$ the output (label)
space. In
this work, we only consider classification tasks, so $\mathcal{Y} = \{1,
..., K\}$. The dataset is
$\mathcal{D} \deq \{(x_{1}, y_{1}), ..., (y_{n}, x_{n})\}$. Denote $x_{k:l} \deq (x_{k}, x_{k+1}, ..., x_{l-1}, x_{l})$. We define a \emph{model} for the supervised learning problem as a
conditional probability distribution $p(y|x)$, namely, a function such
that for each $x\in \mathcal{X}$, $\sum_{y\in\mathcal{Y}} p(y|x)=1$. A \emph{model class}, or
\emph{architecture},  is a set of models
depending on some parameter $\theta$: $\mathcal{M} = \{p_{\theta}, \theta\in \Theta\}$.
The \emph{Kullback--Leibler divergence} between two distributions is
$\kl(\mu\|\nu) = \BE_{X\sim
\mu}[\log_{2}\frac{\mu(x)}{\nu(x)}]$.

\subsection{Models and codelengths}

We recall a basic result of compression theory
\citep{Shannon2001}.

\begin{proposition}[Shannon--Huffman code] 
\label{prop:shannon}
Suppose that Alice and Bob have agreed in advance on a model $p$, and
both know the inputs $x_{1:n}$. Then
there exists a code to transmit the labels $y_{1:n}$ losslessly with codelength
(up to at most one bit on the whole sequence)
\begin{equation}
    \label{eq:shannon}
    L_p(y_{1:n} | x_{1:n}) = -\sum_{i=1}^{n}\log_{2}p(y_{i}|x_{i})
\end{equation}
\end{proposition}

This bound is known to be optimal if the data are independent and
coming from the model $p$ \citep{Mackay}. The one additional bit in the Shannon--Huffman code is incurred only once
for the whole dataset \citep{Mackay}. With large datasets this is
negligible. Thus, from now on we will systematically omit the $+1$ as
well as admit non-integer codelengths \citep{Grunwald}. We will use
the terms \emph{codelength} or \emph{compression bound} interchangeably.

This bound is exactly the categorical \emph{cross-entropy
loss} evaluated on the model $p$. Hence, trying to minimize the
description length of the outputs over the parameters of a model class is equivalent to minimizing the usual classification loss.

Here we do not consider the practical
implementation of compression algorithms: we only care about the theoretical \emph{bit length} of their associated encodings.
We are interested in measuring the amount of information contained in the
data, the mutual information between input and output, and how it is
captured by the model. Thus, we will directly work with codelength functions.

\medskip

An obvious limitation of the bound \eqref{eq:shannon} is that Alice and Bob both have to 
know the model $p$ in advance. This is problematic if the model must be
learned from the data. 

\newcommand{\unif}{\mathrm{unif}}

\subsection{Uniform encoding} 
The uniform distribution $p_\unif(y|x)=\frac{1}{K}$ over the $K$ classes
does not require any learning from the data, thus no additional
information has to be transmitted. Using $p_\unif(y|x)$
\eqref{eq:shannon} yields a
codelength
\begin{equation}
L^\unif(y_{1:n}|x_{1:n})=n\log_2 K
\end{equation}

This \emph{uniform encoding} will be a sanity check against which to
compare the other encodings in this text. 
For MNIST, the uniform encoding cost is $60000\times\log_{2}10 =  199 \kbits$. For CIFAR, the uniform encoding cost is $50000\times\log_{2}10 =  166 \kbits$. 
\subsection{Mutual information between inputs and outputs}
\label{sec:mi}

Intuitively, the only way to beat a trivial encoding of the outputs is to use the
mutual information (in a loose sense) between the inputs and outputs.

This can be formalized as follows. Assume that the inputs and outputs
follow a ``true'' joint distribution $q(x,y)$. Then any transmission
method with
codelength $L$ satisfies \citep{Mackay}
\begin{equation}
\BE_q [L(y|x)] \geq H(y|x)
\end{equation}

Therefore, the gain (per data point) between the codelength $L$
and the trivial codelength $H(y)$ is
\begin{equation}
H(y)-\BE_q [L(y|x)]\leq H(y)-H(y|x)=I(y;x)
\end{equation}
the mutual information between inputs and outputs \citep{Mackay}.

Thus, the gain of \emph{any} codelength compared to the uniform code is limited by
the amount of mutual information between input and output. (This bound is
reached with the true model $q(y|x)$.) Any
successful compression of the labels is, at the same time, a direct estimation of the
mutual information between input and output. The latter is the central quantity in the Information Bottleneck approach to deep learning models \citep{Shwartz-Ziv}.

Note that this still makes sense without assuming a true underlying
probabilistic model, by replacing the mutual information $H(y)-H(y|x)$ with the
``absolute'' mutual information $K(y)-K(y|x)$ based on Kolmogorov
complexity $K$ \citep{Li2008a}.

\section{Compression Bounds via Deep Learning}

 \begin{table}[t]
  \caption{\textbf{Compression bounds via Deep Learning.} Compression
  bounds given by different codes on two datasets, MNIST and CIFAR10. The
  \emph{Codelength} is the number of bits necessary to send the labels 
  to someone who already has the inputs. This codelength
  \emph{includes} the description length of the model. The
  \emph{compression ratio} for a given code is the ratio between its
  codelength and the codelength of the uniform code.
  The \emph{test accuracy} of a model is the accuracy of its predictions
  on the test set. For 2-part and network compression codes, we report
  results from \citep{Han2015} and \citep{Xu2017}, and for the intrinsic dimension code, results from \citep{Li2018}. The values in the table
  for these codelengths and compression ratio are lower bounds, only
  taking into account the codelength of the weights, and not the
  codelength of the data encoded with the model (the final loss is not always
  available in these publications). 
  For variational and  prequential codes, we selected the model and hyperparameters providing the best compression bound.}
  
  \begin{center}
  \fontsize{9pt}{9pt}\selectfont
  \begin{sc}
  \begin{tabular}[h]{lcccccc}
    \toprule
    \multicolumn{1}{l}{\normalsize Code} & \multicolumn{3}{c}{\normalsize mnist} & \multicolumn{3}{c}{\normalsize cifar10}                                                \\
    \cmidrule(lr){2-4} \cmidrule(lr){5-7} 
    {}                                   & Codelength
                                         & Comp.                                 & Test          & Codelength      & Comp.              & Test                            \\
    {}                                   & ($\kbits$)
                                         & Ratio                                 & Acc           & ($\kbits$)      & Ratio
                                         & Acc                                                                                                                            \\
    \midrule
    Uniform                              & 199                                   & 1.            & 10\%            & 166                & 1.            & 10\%            \\
    \midrule
    float32 2-part                       & $> 8.6 \mathrm{Mb}$                   & $> 45.$       & 98.4\%          & $> 428\mathrm{Mb}$ & $> 2500.$     & \textbf{92.9\%} \\
    Network compr.                       & $> 400$                               & $> 2.$        & 98.4\%          & $> 14\mathrm{Mb}$  & $> 83.$       & \textbf{93.3\%} \\
    Intrinsic dim.                       & $> 9.28$                              & $> 0.05$      & 90\%            & $> 92,8$           & $> 0.56$      & 70\%            \\
    \midrule
    Variational                          & 22.2                                  & 0.11          & 98.2\%          & 89.0               & 0.54          & 66,5\%          \\
    Prequential                          & \textbf{4.10}                         & \textbf{0.02} & \textbf{99.5\%} & \textbf{45.3}      & 0.27          & \textbf{93.3\%} \\
    \bottomrule
  \end{tabular}
  \end{sc}
  \end{center}
  
  \vskip -0.1in
  \label{tab:results}
\end{table}

Various compression methods from the MDL toolbox can be used on deep
learning models. (Note that a given model can be stored or encoded in
several ways, some of which may have large codelengths. A good model in
the MDL sense is one that admits at least one good encoding.)

\subsection{Two-Part Encodings} 

Alice and Bob
can first agree on a model class (such as ``neural networks with two
layers and 1,000 neurons per layer''). However, Bob does not have access
to the labels, so Bob cannot train the parameters of the model.
Therefore, if Alice wants to use such a parametric model, the parameters
themselves have to be transmitted. Such codings in which Alice first transmits the parameters of a model,
then encodes the data using this parameter, have been called
\emph{two-part codes} \citep{Grunwald}.

\newcommand{\tp}{\mathrm{2\text{-part}}}
\newcommand{\param}{\mathrm{param}}

\begin{definition}[Two-part codes]
Assume that Alice and Bob have first agreed on a model class $(p_\theta)_{\theta\in\Theta}$. 
Let $L_\param(\theta)$ be any encoding scheme for parameters
$\theta\in\Theta$. 
Let $\theta^\ast$ be any
parameter. The corresponding \emph{two-part codelength} is
\begin{equation}
L^{\tp}_{\theta^\ast}(y_{1:n}|x_{1:n}) \deq L_\param(\theta^\ast)+L_{p_{\theta^\ast}}(y_{1:n}|x_{1:n})
 = L_\param(\theta^\ast)-\sum_{i=1}^n \log_2 p_{\theta^\ast}(y_i|x_i)
\end{equation}
\end{definition}
An obvious possible code 
$L_\param$ for $\theta$ is the
standard float32 binary encoding for $\theta$, for which $L_\param(\theta) = 32 \dim(\theta)$. In deep learning, two-part codes are widely inefficient and much worse
than the uniform encoding \citep{Graves2011}. For a model with 1 million
parameters, the two-part code with float32 binary encoding will amount to
$32\mbits$, or 200 times the uniform encoding on CIFAR10.

\subsection{Network Compression} 
\label{sec:netcompression}
The  practical encoding of trained models is a well-developed research
topic, e.g., for use on small devices such as 
cell
phones. Such encodings can be seen as two-part
codes using a clever code for $\theta$ instead of encoding every
parameter on $32$ bits.
Possible strategies include training a \emph{student layer} to approximate a well-trained network \citep{Ba2013,Romero2015}, or pipelines involving retraining, pruning, and quantization of the model weights \citep{Han2015,Han2015a,Simonyan2014,Louizos, See, Ullrich2017}. 

Still, the resulting codelengths (for compressing the labels given the
data) are way above the uniform compression bound for image
classification (Table~\ref{tab:results}).

Another scheme for network compression, less used in practice but very
informative, is to sample a random low-dimensional affine subspace in
parameter space and to optimize in this subspace
\citep{Li2018}. The number of parameters is thus reduced to the dimension
of the subspace and we can use the associated two-part encoding.
(The random subspace can be transmitted via a pseudorandom
seed.)
Our methodology to derive compression bounds from 
\citep{Li2018} is detailed in
Appendix~\ref{app:intrinsic-dim}.

\subsection{Variational and Bayesian Codes}
\label{sec:varcode}

Another strategy for encoding weights with a limited precision is to
represent these weights by random variables: the uncertainty on $\theta$
represents the precision with which $\theta$ is transmitted.  The
\emph{variational code} turns this into an explicit encoding scheme,
thanks to the
\emph{bits-back} argument \citep{Honkela2004}. Initially a way to compute codelength bounds with
neural networks \citep{Hinton}, this is now often seen as a regularization
technique \citep{Blundell}. This method yields the following codelength.

\begin{definition}[Variational code]
  \label{boundvar}
  Assume that Alice and Bob have agreed on a model class
  $(p_\theta)_{\theta\in\Theta}$ and a prior $\alpha$ over
  $\Theta$. Then for any distribution $\beta$ over $\Theta$, 
  there exists an encoding with codelength 
  \begin{equation}
    \label{eq:boundvar}
    L^{\mathrm{var}}_{\beta}(y_{1:n}|x_{1:n}) =
    \kl\left(\beta\|\alpha\right) + \BE_{\theta \sim
    \beta}\big[L_{p_{\theta}}(y_{1:n}|x_{1:n})\big] =
    \kl\left(\beta\|\alpha\right)- \BE_{\theta \sim
    \beta}\bigg[\sum_{i=1}^{n}\log_2 p_{\theta}(y_{i}|x_{i})\bigg] 
  \end{equation}
\end{definition}

This 
can be minimized over
$\beta$, by choosing a parametric model class
$(\beta_\phi)_{\phi\in\Phi}$, and minimizing \eqref{eq:boundvar} over
$\phi$. A common model class for $\beta$ is the set of multivariate
Gaussian distributions  $\{\mathcal{N}(\mu, \Sigma), \mu\in\BR^{d},
\Sigma \;  \text{ diagonal}\}$, and $\mu$ and $\Sigma$ can be optimized
with a stochastic gradient descent algorithm \citep{Graves2011,
Kucukelbir2017}. $\Sigma$ can be interpreted as the precision with which
the parameters are encoded.

\newcommand{\bayes}{\mathrm{Bayes}}

The variational bound $L^{\mathrm{var}}_{\beta}$ is an upper bound for
the Bayesian description length bound of the Bayesian model $p_\theta$ with
parameter $\theta$ and prior $\alpha$. Considering the
Bayesian distribution of $y$,
%
\begin{equation}
  \label{eq:distrbayes}
  p_\bayes(y_{1:n}|x_{1:n}) = \int_{\theta\in\Theta} p_{\theta}(y_{1:n}|x_{1:n})\alpha(\theta)d\theta, 
\end{equation}
then Proposition~\ref{prop:shannon} provides an associated code
via \eqref{eq:shannon} with model $p_\bayes$:
  $L^\bayes(y_{1:n}|x_{1:n}) = -\log_{2} p_\bayes(y_{1:n}|x_{1:n})$
Then, for any $\beta$ we have \citep{Graves2011}
\begin{equation}
  \label{eq:compbayesvi}
  L^{\mathrm{var}}_{\beta}(y_{1:n}|x_{1:n}) \geq L^\bayes(y_{1:n}|x_{1:n})
\end{equation}
with equality if and only if $\beta$ is equal to the Bayesian posterior
$p_\bayes(\theta|x_{1:n},y_{1:n})$. Variational methods can be used as 
approximate Bayesian inference for intractable Bayesian posteriors.

We computed practical compression bounds with variational methods on
MNIST and CIFAR10. Neural networks that give the best
variational compression bounds appear to be smaller than
networks trained the usual way.
We tested various
fully connected networks and convolutional networks (Appendix
\ref{app:var}): the models that
gave the best variational compression bounds were small LeNet-like networks.
To test the link between compression and test accuracy, in Table
\ref{tab:results} we report
the best model based
on compression, not test accuracy.
This results in a drop of test accuracy with respect to other settings.

On MNIST, this provides a codelength of the labels (knowing the
inputs) of $24.1\kbits$, i.e., a compression ratio of
$0.12$. The corresponding model achieved $95.5\%$
accuracy on the test set.

On CIFAR, we obtained a codelength of $89.0\kbits$, i.e., a compression
ratio of $0.54$. The corresponding model achieved $61.6\%$ classification accuracy on the test set.

We can make two observations. First, choosing the model class which
minimizes variational codelength selects smaller deep learning models
than would cross-validation. Second, the
model with best variational codelength has low classification accuracy on the test set on
MNIST and CIFAR, compared to models trained in a non-variational way.
This aligns with a common criticism 
of Bayesian 
methods as too conservative for model selection
compared with cross-validation
\citep{Rissanen1992,Foster1994,Barron1999,Grunwald}.

\subsection{Prequential or Online Code}
\label{sec:online-learning}

\newcommand{\preq}{\mathrm{preq}}

The next coding procedure shows that deep neural models
which generalize well also compress well.

The prequential (or online) code is a way to encode both the model and
the labels without \emph{directly} encoding the weights, based on the
\emph{prequential approach to statistics} \citep{Dawid1984}, by using
\emph{prediction strategies}. Intuitively, a model with default values is
used to encode the first few data; then the model is trained on these
few encoded data; this partially trained model is used to encode the next
data; then the model is retrained on all data encoded so far; and so
on.

Precisely, we call $p$ a \emph{prediction strategy} for
predicting the labels in $\mathcal{Y}$ knowing the inputs in
$\mathcal{X}$ if for all $k$, $p(y_{k+1}|x_{1:k+1}, y_{1:k})$ is a
conditional model; namely, any strategy for predicting the $k+1$- label
after already having seen $k$ input-output pairs. In particular, such a
model may \emph{learn} from the first $k$
data samples. Any prediction strategy $p$ defines a model on the whole
dataset:
\begin{equation}
  p^\preq(y_{1:n}|x_{1:n}) = p(y_{1}|x_{1})\cdot p(y_{2}|x_{1:2}, y_{1})\cdot \ldots\cdot p(y_{n}|x_{1:n}, y_{1:n-1})
\end{equation}

Let $(p_{\theta})_{\theta\in\Theta}$ be a deep learning model. We assume
that we have a learning algorithm which computes, from any number of data
samples $(x_{1:k},y_{1:k})$, a trained parameter vector $\hat\theta(x_{1:k},y_{1:k})$.
Then the data is encoded in an incremental way: at each step $k$, 
$\hat\theta(x_{1:k},y_{1:k})$ is used to predict $y_{k+1}$.

In practice, the learning procedure $\hat\theta$ may only reset and
retrain the network at certain timesteps. We choose timesteps $1 = t_{0}
< t_{1} < ... < t_{S} = n$, and we encode the data by blocks, always
using the model learned from the already transmitted data
(Algorithm~\ref{alg:preq} in Appendix~\ref{app:preq}). 
A uniform encoding is used for the
first few points.
(Even though the encoding
procedure is called ``online'', it does not mean that only the most
recent sample is used to update the parameter $\hat\theta$:  the
optimization procedure $\hat{\theta}$ can be any
predefined technique using all the previous samples
$(x_{1:k}, y_{1:k})$, only requiring that the algorithm has an explicit
stopping criterion.)
This yields the following description length:


\begin{definition}[Prequential code]
Given a model $p_\theta$,
a learning algorithm $\hat{\theta}(x_{1:k},y_{1:k})$, and retraining
timesteps $1 = t_{0}
< t_{1} < ... < t_{S} = n$,
the \emph{prequential} codelength is
  \label{boundpreq}
  \begin{equation}
    \label{eq:boundpreq}
    L^\preq(y_{1:n}|x_{1:n}) = t_{1}\log_{2}K + \sum_{s=0}^{S-1}-\log_{2} p_{\hat\theta_{t_{s}}}(y_{t_{s}+1:t_{s+1}}|x_{t_{s}+1:t_{s+1}})
  \end{equation}
where for each $s$, $\hat\theta_{t_{s}}=\hat\theta(x_{1:t_s},y_{1:t_s})$
is the parameter learned on data samples $1$ to $t_s$.
\end{definition}

The model parameters are never encoded explicitly in this method. The
difference between the prequential codelength $L^\preq(y_{1:n}|x_{1:n})$
and the log-loss $\sum_{t=1}^n -\log_2 p_{\hat\theta_{t_K}} (y_t|x_t)$ of
the final trained model, can be interpreted as the amount of information
that the trained parameters contain about the data contained: the former
is the data codelength if Bob does not know the parameters, while the latter is
the codelength of the same data knowing the parameters.

Prequential codes depend on the performance of the underlying training
algorithm, and take advantage of the model's generalization ability from
the previous data to the next. In particular, the model training should yield good
generalization performance from data $[1;t_s]$ to data $[t_s+1;t_{s+1}]$.

In practice, optimization procedures for neural networks may be stochastic
(initial values, dropout, data augmentation...), and Alice and Bob need
to make all the same random actions in order to get the same final model.
A possibility is to 
agree on a random seed $\omega$ (or pseudorandom numbers) beforehand,
so that the random optimization procedure
$\hat\theta(x_{1:t_{s}}, y_{1:t_{s}})$ is deterministic given $\omega$,
Hyperparameters may also be transmitted first (the
cost of sending a few numbers is small).

Prequential coding with deep models provides excellent compression
bounds.
On MNIST, we computed the description length of the labels with different
networks (Appendix~\ref{app:preq}). The best compression bound was given
by a convolutional network of depth 8. It achieved a description length
of $4.10\kbits$, i.e., a compression ratio of $0.021$, with $99.5\%$ test
set accuracy (Table~\ref{tab:results}). This codelength is 6 times smaller than the variational
codelength.

\begin{figure}[t]
    \includegraphics[width=\linewidth]{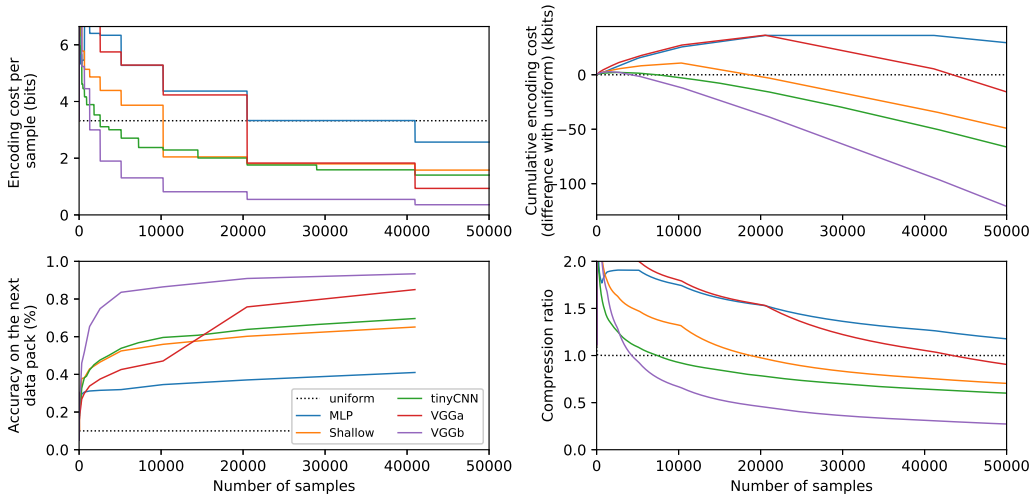}
    \caption{\textbf{Prequential code results on CIFAR.}
    Results of prequential encoding on CIFAR with 5 different models: a
    small Multilayer Perceptron (MLP), a shallow network, a small convolutional layer
    (tinyCNN), a VGG-like network without data augmentation and batch
normalization (VGGa) and the same VGG-like architecture with data
    augmentation and batch normalization (VGGb) (see Appendix~\ref{app:preq}). Performance
    is reported during online training, as a function of the number of
    samples seen so far. Top left:  codelength per sample
    (log loss) on a pack of data $[t_k;t_{k+1})$ given data $[1;t_k)$. Bottom left: test accuracy on a pack of data
    $[t_k;t_{k+1})$ given data $[1;t_k)$, as a function of $t_k$.
    Top right: difference between the prequential cumulated codelength on data $[1;t_k]$, and
    the uniform encoding. Bottom right: compression ratio of the prequential
    code on data $[1;t_k]$.  }
    \label{fig:cifarcomp}
  \end{figure}

On CIFAR, we tested
a simple multilayer perceptron, a shallow network, a small convolutional
network, and a VGG convolutional network \citep{Simonyan2014} first
without data augmentation or batch normalization (VGGa)
\citep{Ioffe2015}, then with both of them (VGGb) (Appendix~\ref{app:preq}). The results are in
Figure \ref{fig:cifarcomp}.
The best compression bound was obtained with
VGGb, achieving a codelength of $45.3\kbits$, i.e., 
a compression ratio of $0.27$, and $93\%$ test set accuracy
(Table~\ref{tab:results}).
This codelength is twice smaller than the variational codelength. The
difference between VGGa and VGGb also shows the impact of the training
procedure on codelengths for a given architecture.

\paragraph{Model Switching.} A weakness of prequential codes is the \emph{catch-up phenomenon}
\citep{VanErven2011}. Large architectures might overfit during the first
steps of the prequential encoding, when the model is trained with few
data samples.  Thus the encoding cost of the first packs of data
might be worse than with the uniform code. Even after the encoding cost
on current labels becomes lower, the cumulated codelength may need a
lot of time to ``catch up'' on its initial lag.
\optional{This can be observed in practice
with neural networks:} in Fig.~\ref{fig:cifarcomp}, the VGGb model
needs 5,000 samples on CIFAR to reach a cumulative compression ratio $<\!1$, even
though the encoding cost per label becomes drops below uniform after just 1,000
samples. This is efficiently solved by \emph{switching}
\citep{VanErven2011} between models (see
Appendix~\ref{sec:extend-class-models}). Switching further improves the practical
compression bounds\optional{, even when just switching between copies of the same
model with different SGD stopping times} (Fig.~\ref{fig:selfswitch},
Table \ref{tab:resultsswitch}).

\section{Discussion}
\label{sec:discussions}


\paragraph{Too Many Parameters in Deep Learning Models?}
>From an information theory perspective, the goal of a model is to extract
as much mutual information between the labels and inputs as
possible—equivalently (Section~\ref{sec:mi}), to compress the labels.
This cannot be achieved with 2-part codes or practical network
compression. With the variational code, the models do compress the data,
but with a worse prediction performance: one could conclude that deep
learning models that achieve the best prediction performance cannot
compress the data. 


Thanks to the prequential code, we have seen that deep learning models,
even with a large number of parameters, compress the data well: from an
information theory point of view, \emph{the number of parameters is not
an obstacle to compression}. This is consistent with Chaitin's hypothesis
that ``\emph{comprehension is compression}'', contrary to previous
observations with the variational code.

 
\paragraph{Prequential Code and Generalization.}
The prequential encoding shows that a model that generalizes well for
every dataset size, will
compress well.
The
efficiency of the prequential code is directly due to the generalization
ability of the model at each time.

Theoretically, three of the codes (two-parts, Bayesian, and
prequential based on a maximum likelihood or MAP estimator) are known to be asymptotically equivalent under strong
assumptions
($d$-dimensional \emph{identifiable} model,
data coming from the model, 
 suitable Bayesian prior, and technical
assumptions ensuring the effective dimension of the trained model is not
lower than $d$):
in that case, these three methods yield a codelength
$L(y_{1:n}|x_{1:n}) = nH(Y|X) + \frac{d}{2} \log_{2}n + \mathcal{O}(1)$
\citep{Grunwald}. This corresponds to
the BIC criterion for model selection. 
Hence there was no obvious reason for the prequential code to be an order of
magnitude better than the others.

However, deep learning models do not
usually satisfy \emph{any} of these hypotheses.
Moreover, our prequential codes are not based on the maximum
likelihood estimator at each step, but on standard deep learning methods
(so training is regularized at least by
dropout and early stopping).

\paragraph{Inefficiency of Variational Models for Deep Networks.}

The objective of variational methods is equivalent to minimizing a
description length. Thus, on our image classification tasks, variational methods
do not have good results \emph{even for their own objective}, compared to
prequential codes. This makes their relatively poor results at test time
less surprising.

Understanding this observed inefficiency of variational methods 
is an open problem. As stated in
\eqref{eq:compbayesvi}, the variational codelength is an upper bound for
the Bayesian codelength. More precisely,
\begin{equation}
\label{eq:varbayes}
    L_\beta^{\mathrm{var}}(y_{1:n}|x_{1:n}) = L^{\mathrm{Bayes}}(y_{1:n}|x_{1:n}) + \kl\left(p_\bayes(\theta|x_{1:n}, y_{1:n})\|\beta\right)
\end{equation}
with notation as above, and with
$p_\bayes(\theta|x_{1:n}, y_{1:n})$ the Bayesian posterior on $\theta$
given the data.
Empirically, on MNIST and CIFAR, we observe that
$
  L^\preq(y_{1:n}|x_{1:n}) \ll L_\beta^{\mathrm{var}}(y_{1:n}|x_{1:n}).
$

Several phenomena could contribute to this gap. First, the optimization
of the parameters $\phi$ of the approximate
Bayesian posterior might be imperfect. Second, even
the optimal distribution $\beta^{*}$ in the variational class might not
approximate the posterior $p_\bayes(\theta|x_{1:n}, y_{1:n})$ well,
leading to a large $\kl$ term in \eqref{eq:varbayes}; this would be a
problem with the choice of variational posterior class $\beta$.
On the other hand we do not expect the choice of Bayesian prior to be a key factor: we
tested Gaussian priors with various variances as well as a conjugate
Gaussian prior, with similar results.
Moreover, Gaussian initializations
and L2 weight decay (acting like a Gaussian prior) are common in deep
learning.
Finally,
the (untractable) Bayesian codelength based on the exact posterior might
itself be larger than the prequential codelength. This would be a problem
of underfitting with parametric Bayesian inference, perhaps related to
the catch-up phenomenon or to the known conservatism of Bayesian model
selection (end of Section~\ref{sec:varcode}).

\section{Conclusion}
\label{sec:conclusion}

Deep learning models can represent the data \emph{together with the
model} in fewer bits than a naive encoding, despite their many
parameters.  However, we were surprised to observe that variational
inference, though explicitly designed to minimize such codelengths,
provides very poor such values compared to a simple incremental coding
scheme. Understanding this limitation of variational inference  is a topic for future research.

\section*{Acknowledgments}
First, we would like to thank the reviewers for their careful reading and their questions and comments. We would also like to thank Corentin Tallec for his technical help, and David Lopez-Paz, Moustapha Cissé, Gaétan Marceau Caron and Jonathan Laurent for their helpful comments and advice.

\todo{Acknowledgments: DLP, Corentin, Moustapha, Gaétan.}

\bibliography{library}

\vfill

\pagebreak
\appendix

\section{Fake labels are not compressible}
\label{app:fake-noncompressible}

In the introduction, we stated that fake labels could not be compressed.
This means that the optimal codelength for this labels is \emph{almost}
the uniform one. This can be formalized as follows. We define a
\emph{code} for $y_{1:n}$ as any program (in a reference Turing machine)
that outputs $y_{1:n}$, and denote $L(y_{1:n})$ the length of this
program, or $L(y_{1:n}|x_{1:n})$ for programs that may use $x_{1:n}$ as
their input.

\begin{proposition} 
\label{prop:non-compressible}
Assume that $x_{1}, ..., x_{n}$ are inputs, and that $Y_{1}, ..., Y_{n}$
are iid random labels uniformly sampled in $\{1, ..., K\}$. Then for any
$\delta \in \BN^{*}$, with probability $1 - 2^{-\delta}$ the values
$Y_1,\ldots,Y_n$ satisfy that
 for any possible
coding procedure $L$ (even depending on the values of $x_{1:n}$), the codelength of $Y_{1:n}$ is at least
\begin{align}
  L(Y_{1:n}|x_{1:n}) &\geq nH(Y) - \delta - 1\\
                     &= n\log_{2}K - \delta - 1.
\end{align}
\end{proposition}

We insist that this does not require any assumptions on the coding
procedure used, so this result holds for all possible models. Moreover,
this is really a property of the sampled values $Y_1,\ldots Y_n$: most
values of $Y_{1:n}$ can just not be compressed by any algorithm.

\begin{proof}
  This proposition is a standard counting argument, or an immediate consequence of Theorem 2.2.1 in \citep{Li2008a}.
  Let $\mathcal{A} = \{1, ..., K\}^{n}$ be the set of all possible outcomes for the sequence of random labels. We have $|\mathcal{A}| = K^{n}$.
  Let $\delta$ be an integer, $\delta \in \BN^{*}$, we want to know how many elements in $\mathcal{A}$ can be encoded in less than $\log_{2}|\mathcal{A}| - \delta$ bits. We consider, on a given Turing machine, the number of programs of length less than $\lfloor\log_{2}|\mathcal{A}| - \delta\rfloor$. This number is less than :
  \begin{align}
    \sum_{i=0}^{\lfloor\log_{2}|\mathcal{A}|\rfloor - \delta - 1}2^{i} &= 2^{\lfloor\log_{2}|\mathcal{A}|\rfloor - \delta} - 1 \\
    &\leq 2^{-\delta}|\mathcal{A}| - 1
  \end{align}
  Therefore, the number of elements in $\mathcal{A}$ which can be
  described in less than $\log_{2}|\mathcal{A}| - \delta$ bits is less
  than $2^{-\delta}|\mathcal{A}| - 1$. We can deduce from this that the
  number of elements in $\mathcal{A}$ which cannot be described by
  \emph{any} program in less
  than $2^{-\delta}|\mathcal{A}| - 1$ bits is at least $|\mathcal{A}|(1 -
  2^{-\delta})$. Equivalently, there are at least
  $|\mathcal{A}|(1 - 2^{-\delta})$ elements $(y_{1}, ..., y_{n})$ in
  $|\mathcal{A}|$ such that for any coding scheme, $L(y_{1:n}|x_{1:n}) \geq n\log_{2}K - \delta
  - 1$. Since the random labels $Y_{1}, ..., Y_{n}$ are uniformly
  distributed, the result follows.
\end{proof}

\section{Technical details on compression bounds with random affine subspaces}
\label{app:intrinsic-dim}

We describe in Algorithm~\ref{alg:intr-dim} the detailed procedure which
allows to compute compression bounds with the random affine subspace
method \citep{Li2018}. To compute the numerical results in
Table~\ref{tab:results}, we took the \emph{intrinsic dimension} computed
in the original paper, and considered that the precision of the parameter
was 32 bits, following the authors' suggestion. Then, the description
length of the model itself is $32 \times$ the intrinsic dimension. This
does not take into account the description length of the labels given the
model, which is non-negligible (to take this quantity into account, we
would need to know the loss on the training set of the model, which was
not specified in the original paper). Thus we only get a lower bound.

\begin{algorithm}[h]
   \caption{Encoding with random affine subspaces}
   \label{alg:intr-dim}
\begin{algorithmic}
  \STATE Alice transmits a parametric model $(p_{\theta})_{\theta\in\Theta}$.
   \STATE Alice transmits the random seed $\omega$ (if using stochastic
   optimization), and a dimension $k$.
   \STATE Alice and Bob both sample a random affine subspace $\tilde \Theta \subset \Theta$, with the seed $\omega$. This means that they sample $\theta_{0}$ and a matrix $W$ of dimension $k\times d$ where $d$ is the dimension of $\Theta$. It defines a new parametric model $\tilde p_{\phi} = p_{\theta_{0} + W\cdot \phi}$
   \STATE Alice optimizes the parameter $\phi^{*}$ with a gradient descent algoritm in order to minimize $-\log_{2}\tilde p_{\phi}(y_{1:n}|x_{1:n})$.
   \STATE Alice sends $\phi^{*}$ with a precision $\varepsilon$ to Bob. It costs $k\times\log_{2}\varepsilon$.
   \STATE Alice sends the labels $y_{1:n}$ with the models $\tilde p_{\phi^{*}}$. It costs $-\log_{2}\tilde p_{\phi^{*}}(y_{1:n}|x_{1:n})$
\end{algorithmic}
\end{algorithm}

For MNIST, the model with the smaller intrinsic dimension is the LeNet,
which has an intrinsic dimension of 290 for an accuracy of $90\%$ (the
threshold at which \citep{Li2018} stop by definition, hence the
performance in Table~\ref{tab:results}). This
leads to a description length for the model of $9280$ bits, which
corresponds to a $0.05$ compression ratio, without taking into account
the description length of the labels given the model.

For CIFAR, again with the LeNet architecture, the intrinsic dimension is 2,900. This leads to a description length for the model of $92800$ bits, which corresponds to a $0.05$ compression ratio, without taking into account the description length of the labels given the model.

These bounds could be improved by optimizing the precision $\varepsilon$.
Indeed, reducing the precision makes the model less accurate and
increases the encoding cost of the labels with the model, but it
decreases the encoding cost of the parameters. Therefore, we could find
an optimal precision $\varepsilon^{*}$ to improve the compression bound.
This would be a topic for future work.

\section{Technical Details on Variational Learning for
Section~\ref{sec:varcode}}
\label{app:var}

Variational learning was performed using the library Pyvarinf \citep{Tallec2018}.

We used a prior $\alpha = \mathcal{N}(0, \sigma_{0}^{2}I_{d})$ with $\sigma_{0} = 0.05$, chosen to optimize the compression bounds.

The chosen class of posterior was the class of multivariate gaussian distributions with diagonal covariance matrix $\{\mathcal{N}(\mu, \Sigma)\; ,\;  \mu\in\BR^{d} \; \Sigma \; \mathrm{ diagonal}\}$. It was parametrized by $(\beta_{\mu, \rho})_{(\mu, \rho) \in \BR^{d}\times\BR^{d}}$, with $\sigma \in \BR^{d}$ defined as $\sigma_{i} = \log(1+\exp({\rho_{i}}))$, and the covariance matrix $\Sigma$ as the diagonal matrix with diagonal values $\sigma_{1}^{2}, ..., \sigma_{d}^{2}$.

We optimize the bound~\eqref{eq:boundvar} as a function of ($\mu, \rho$) with a gradient descent method, and estimate its values and gradient with a Monte-Carlo method.
Since the prior and posteriors are gaussian, we have an explicit formula for the first part of the variational loss $\kl(\beta_{\mu, \rho}\|\alpha)$ \citep{Hinton}. Therefore, we can easily compute its values and gradients. 
For the second part 
\begin{equation}
(\mu, \rho) \rightarrow \BE_{\theta \sim
    \beta_{\mu, \rho}}\bigg[\sum_{i=1}^{n}-\log_2 p_{\theta}(y_{i}|x_{i})\bigg],
\end{equation}
we can use the following proposition \citep{Graves2011}. For any function
$f \colon \Theta \rightarrow \BR$, we have
\begin{align}
  \frac{\partial}{\partial\mu_{i}}\BE_{\theta\sim\beta_{\mu, \rho}}[f(\theta)] &= \BE_{\theta\sim\beta_{\mu, \rho}}\Big[\frac{\partial f}{\partial\theta_{i}} (\theta)\Big] \\
\frac{\partial}{\partial\rho_{i}}\BE_{\theta\sim\beta_{\mu, \rho}}[f(\theta)] &= \frac{\partial \sigma_{i}}{\partial\rho_{i}}\cdot \BE_{\theta\sim\beta_{\mu, \rho}}\Big[\frac{\partial f}{\partial\theta_{i}} \cdot \frac{\theta_{i} - \mu_{i}}{\sigma_{i}}\Big] 
\end{align}
Therefore, we can estimate the values and gradients of~\eqref{eq:boundvar}  with a Monte-Carlo algorithm:
\begin{align}
  \frac{\partial}{\partial\mu_{i}}\BE_{\theta\sim\beta_{\mu, \rho}}[f(\theta)] &\approx \sum_{s=1}^{S}\frac{\partial f}{\partial\theta_{i}} (\theta^{s}) \\
\frac{\partial}{\partial\rho_{i}}\BE_{\theta\sim\beta_{\mu, \rho}}[f(\theta)] &\approx \frac{\partial \sigma_{i}}{\partial\rho_{i}}\cdot \sum_{s=1}^{S}\frac{\partial f}{\partial\theta_{i}}(\theta^{s}) \cdot \frac{\theta^{s}_{i} - \mu_{i}}{\sigma_{i}}
\end{align}
where $\theta^{1}, ..., \theta^{S}$ are sampled from $\beta_{\mu, \rho}$. In practice, we used $S=1$ both for the computations of the variational loss and its gradients.

We used both convolutional and fully connected architectures, but in our experiments fully connected models were better for compression. For CIFAR and MNIST, we used fully connected networks with two hidden layers of width 256, trained with SGD, with a $0.005$ learning rate and mini-batchs of size 128.

For CIFAR and MNIST, we used a LeNet-like network with 2 convolutional layers with 6 and 16 filters, both with kernels of size 5 and 3 fully connected layers. Each convolutional is followed by a ReLU activation and a max-pooling layer. The code will be publicly available. The first and the second fully connected layers are of dimension 120 and 84 and are followed by ReLU activations. The last one is followed by a softmax activation layer. The code for all models will be publicly available.

During the test phase, we sampled parameters $\hat\theta$ from the learned distribution $\beta$, and used the model $p_{\hat\theta}$ for prediction. This explains why our test accuracy on MNIST is lower than other numerical results \citep{Blundell}, since they use for prediction the averaged model with parameters $\hat\theta = \BE_{\theta\sim\beta_{m, r}}[\theta] = \mu$. But our goal was not to get the best prediction score, but to evaluate the model which was used for compression on the test set.

\section{Technical details on prequential learning}
\label{app:preq}

\paragraph{Prequential Learning on MNIST.}
On MNIST, we used three different models: 
\begin{enumerate}
\item The uniform probability over the labels.
\item A fully connected network or Multilayer Perceptron (MLP) with two hidden layers of dimension 256.
\item A VGG-like convolutional network with 8 convolutional layers with 32, 32, 64, 64, 128, 128, 256 and 256 filters respectively and max pooling operators every two convolutional layers, followed by two fully connected layers of size 256. 
\end{enumerate}
For the two neural networks we used Dropout with probability $0.5$ between the fully connected layers, and optimized the network with the Adam algorithm with learning rate $0.001$.

The successive timestep for the prequential learning $t_{1}, t_{2}, ..., t_{s}$ are 8, 16, 32, 64, 128, 256, 512, 1024, 2048, 4096, 8192, 16384 and 32768.

For the prequential code results in Table \ref{tab:results}, we selected the best model, which was the VGG-like network.


\paragraph{Prequential Learning on CIFAR.}
On CIFAR, we used five different models: 
\begin{enumerate}
\item The uniform probability over the labels.
\item A fully connected network or Multilayer Perceptron (MLP) with two hidden layers of dimension 512.
\item A shallow network, with one hidden layer and width 5000.
\item A convolutional network (tinyCNN) with four convolutional layers with 32 filters, and a maxpooling operator after every two convolutional layers. Then, two fully connected layers of dimension 256. We used Dropout with probability $0.5$ between the fully connected layers.
\item A VGG-like network with 13 convolutional layers from \citep{Zagoruyko2015}. We trained this architecture with two learning procedures. The first one (VGGa) without batch-normalization and data augmentation, and the second one (VGGb) with both of them, as introduced in \citep{Zagoruyko2015}. In both of them, we used dropout regularization with parameter 0.5.
\end{enumerate}

We optimized the network with the Adam algorithm with learning rate 0.001. 

For prequential learning, the timesteps $t_{1}, t_{2}, ..., t_{s}$ were: 10, 20, 40, 80, 160, 320, 640, 1280, 2560, 5120, 10240, 20480, 40960. The training results can be seen in Figure~\ref{fig:cifarcomp}.

For the prequential code, all the results are in Figure \ref{fig:cifarcomp}. For the results in Table \ref{tab:results}, we selected the best model for the prequential code, which was VGGb.

\begin{algorithm}[tb]
   \caption{Prequential encoding}
   \label{alg:preq}
\begin{algorithmic}
   \STATE {\bfseries Input:} data $x_{1:n}, y_{1:n}$, timesteps $1 = t_{0} < t_{1} < ... < t_{S} = n$
   \STATE Alice transmits the random seed $\omega$ (if using stochastic
   optimization).
   \STATE Alice encodes $y_{1:t_{1}}$ with the uniform code. This costs
   $t_{1}\log_{2}K$ bits.
   Bob decodes $y_{1:t_{1}}$.
   \FOR{$s=1$ {\bfseries to} $S-1$}
   \STATE Alice and Bob both compute $\hat\theta_{s} = \hat\theta(x_{1:t_{s}}, y_{1:t_{s}}, \omega)$. 
   \STATE  Alice encodes $y_{t_{s}+1:t_{s+1}}$ with model
   $p_{\hat\theta_{s}}$. This costs $-\log_{2} p_{\hat\theta_{s}}(y_{t_{s}+1:t_{s+1}}|x_{t_{s}+1:t_{s+1}})$ bits
   \STATE Bob decodes $y_{t_{s}+1:t_{s+1}}$
   \ENDFOR
\end{algorithmic}
\end{algorithm}

\section{Switching between models against the \emph{catch-up phenomenon}}
\label{sec:extend-class-models}

\begin{table}[t]
  \caption{\textbf{Compression bounds by switching between models.} Compression
  bounds given by different codes on two datasets, MNIST and CIFAR10. The
  \emph{Codelength} is the number of bits necessary to send the labels 
  to someone who already has the inputs. This codelength
  \emph{includes} the description length of the model. The
  \emph{compression ratio} for a given code is the ratio between its
  codelength and the codelength of the uniform code.
  The \emph{test accuracy} of a model is the accuracy of its predictions
  on the test set.
  For variational and  prequential codes, we selected the model and hyperparameters providing the best compression bound.}
  
  \begin{center}
  \fontsize{9pt}{9pt}\selectfont
  \begin{sc}
  \begin{tabular}[h]{lcccccc}
    \toprule
    \multicolumn{1}{l}{\normalsize Code} & \multicolumn{3}{c}{\normalsize mnist} & \multicolumn{3}{c}{\normalsize cifar10}                                            \\
    \cmidrule(lr){2-4} \cmidrule(lr){5-7} 
    {}                                   & Codelength
                                         & Comp.                                 & Test          & Codelength      & Comp.          & Test                            \\
    {}                                   & ($\kbits$)
                                         & Ratio                                 & Acc           & ($\kbits$)      & Ratio
                                         & Acc                                                                                                                        \\
    \midrule
    Uniform                              & 199                                   & 1.            & 10\%            & 166            & 1.            & 10\%            \\
    \midrule
    Variational                          & 24.1                                  & 0.12          & 95.5\%          & 89.0           & 0.54          & 61,6\%          \\
    Prequential                          & \textbf{4.10}                         & \textbf{0.02} & \textbf{99.5\%} & 45.3           & 0.27          & \textbf{93.3\%} \\
    Switch                               & \textbf{4.05}                         & \textbf{0.02} & \textbf{99.5\%} & \textbf{34.6 } & \textbf{0.21} & \textbf{93.3\%} \\
    Self-Switch                          & \textbf{4.05}                         & \textbf{0.02} & \textbf{99.5\%} & \textbf{34.9}  & \textbf{0.21} & \textbf{93.3\%} \\
    \bottomrule
  \end{tabular}
  \end{sc}
  \end{center}
  
  \vskip -0.1in
  \label{tab:resultsswitch}
\end{table}


\subsection{Switching between model classes}

The solution introduced by \citep{VanErven2011} against the catch-up
phenomenon described in Section~\ref{sec:online-learning}, is to \emph{switch}
between models, to always encode a data block with the best model at that
point. That way, the encoding adapts itself to the number of data samples seen. 
The switching pattern itself has to be encoded.

Assume that Alice and Bob have agreed on a set of prediction strategies
$\mathcal{M} = \{p^{k}, k \in \mathcal{I}\}$. We define the set of switch sequences, $\BS = \{((t_{1}, k_{1}), ..., (t_{L}, k_{L})), 1 = t_{1} < t_{2} < ... < t_{L}\; , \;  k \in \mathcal{I}\}$. 

Let $s = ((t_{1}, k_{1}), ..., (t_{L}, k_{L}))$ be a switch sequence. The
associated prediction strategy $p_{s}(y_{1:n}|x_{1:n})$ uses model
$p^{k_i}$ on the time interval $[t_i;t_{i+1})$, namely
\begin{align}
  \label{eq:defswitch}
  p_{s}(y_{1:i+1}|x_{1:i+1},y_{1:i}) =  p^{K_{i}}(y_{i+1}|x_{1:i+1},y_{1:i})
\end{align}
where $K_{i}$ is such that $K_{i} = k_{l}$ for $t_{l} \leq i < t_{l+1}$.
Fix a prior distribution $\pi$ over switching sequences (see
\citep{VanErven2011} for typical examples).

\newcommand{\switch}{\mathrm{sw}}

\begin{definition}[Switch code]
\label{boundswitch}
Assume that Alice and Bob have agreed on a set of prediction strategies
$\mathcal{M}$ and a prior $\pi$ over $\BS$. The \emph{switch code} first
encodes a switch sequence $s$ strategy, then uses the prequential code with this strategy:
\begin{equation}
  L_s^\switch(y_{1:n},x_{1:n}) = L_{\pi}(s) + L^{\mathrm{preq}}_{p_{s}}(y_{1:n},x_{1:n})  
         = - \log_{2}\pi(s) - \sum_{i=1}^{n}\log_{2}p^{K_{i}}(y_{i}|x_{1:i},y_{1:i-1})
\end{equation}
where $K_{i}$ is the model used by switch sequence $s$ at time $i$.
\end{definition}


We then choose the switching strategy $s^{*}$ wich minimizes $L_s^\switch(y_{1:n},x_{1:n})$.
We tested switching between the uniform model, a small convolutional
network (tinyCNN), and a VGG-like network with two training methods (VGGa, VGGb) (Appendix~\ref{app:preq}).
On MNIST, switching between models does not make much difference. On
CIFAR10, switching by taking the best model on each interval
$[t_k;t_{k+1})$ saves more than $11\kbits$, reaching a
codelength of $34.6\kbits$, and a compression ratio of $0.21$. The cost
$L_\pi(s)$ of encoding the switch $s$ is negligible (see Table~\ref{tab:resultsswitch}).

\subsection{Self-Switch: Switching between variants of a model or
hyperparameters}

\begin{figure}[t]
    \includegraphics[width=\linewidth]{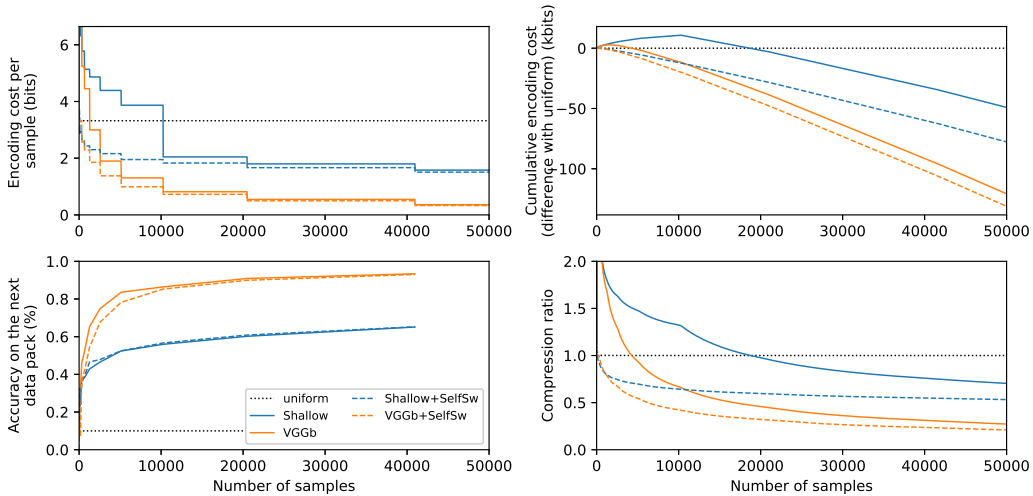}
    \caption{\textbf{Compression with the self-switch method:}
    Results of the self-switch code on CIFAR with 2 different models: the shallow network, and the
    VGG-like network trained with data augmentation and batch normalization (VGGb). Performance
    is reported during online training, as a function of the number of
    samples seen so far. Top: test accuracy on a pack of data
    $[t_k;t_{k+1})$ given data $[1;t_k)$, as a function of $t_k$.
    Second: codelength per sample
    (log loss) on a pack of data $[t_k;t_{k+1})$ given data $[1;t_k)$.
    Third: difference between the prequential cumulated codelength on data $[1;t_k]$, and
    the uniform encoding. Bottom: compression ratio of the prequential
    code on data $[1;t_k]$.  The catch-up phenomenon is clearly visible for both models: even if models with and without the self-switch have similar performances after a training on the entire dataset, the standard model has lower performances than the uniform model (for the 1280 first labels for the VGGb network, and for the 10,000 first labels for the shallow network), and the code length for these first labels is large. The self-switch method solves this problem.}
    \label{fig:selfswitch}
  \end{figure}

With standard switch, it may be cumbersome to work with different models in parallel. Instead,
for models learned by gradient descent, we may use the same architecture
but with different parameter values corresponding obtained at different gradient
descent stopping times.
This is a form of regularization via early stopping.

Let $(p_{\theta})_{\theta\in\Theta}$ be a model class. Let
$\hat\theta_{j}(x_{1:k},y_{1:k})$ be the parameter obtained by some
optimization procedure after $j$ epochs of training on data $[1;k]$. For instance, $j=0$
would correspond to using an untrained model (usually close to the
uniform model).

We call \emph{self-switch code} the switch code obtained by switching
among the family of models with different gradient descent stopping times
$j$ (based on the same parametric family
$(p_{\theta})_{\theta\in\Theta}$).
In practice, this means that at each step of the prequential encoding,
after having seen data $[1;t_k)$, we train the model on those data and
record, at each epoch $j$, the loss obtained on data $[t_k;t_{k+1})$. We then
switch optimally between those. We incur the small additional cost of encoding
the best number of epochs to be used (which was limited to $10$) at each
step.

The catch-up phenomenon and the beneficial effect of the self-switch code can be seen in Figure~\ref{fig:selfswitch}.
%
%


The self-switch code achieves similar compression bounds to the switch
code, while storing only one network. On MNIST, there is no observable
difference. On CIFAR, self-switch is only 300 bits (0.006 bit/label)
worse than
full $4$-architecture switch.

\end{document}